\documentclass[runningheads]{llncs}
\usepackage{graphicx}
\usepackage{cite}
\usepackage{amsmath}
\usepackage{bm}    
\usepackage{varioref}     
\usepackage{float} 

\usepackage{array}
\newcolumntype{P}[1]{>{\centering\arraybackslash}p{#1}}

\usepackage[shortlabels]{enumitem}

\begin{document}

\title{Learning Uncertainty with Artificial Neural Networks for Improved Remaining Time Prediction of Business Processes}
\titlerunning {Uncertainty with Neural Networks in Process Remaining Time Prediction}
%
%
\author{Hans Weytjens\and
Jochen De Weerdt}
\authorrunning{H. Weytjens, J. De Weerdt}
%
\institute{Research Centre for Information Systems Engineering (LIRIS),\\ KU Leuven, Leuven, Belgium\\
\email{\{hans.weytjens,jochen.deweerdt\}@kuleuven.be}}
\maketitle              
\begin{abstract}
Artificial neural networks will always make a prediction, even when completely uncertain and regardless of the consequences. This obliviousness of uncertainty is a major obstacle towards their adoption in practice. Techniques exist, however, to estimate the two major types of uncertainty: model uncertainty and observation noise in the data. Bayesian neural networks are theoretically well-founded models that can learn the model uncertainty of their predictions. Minor modifications to these models and their loss functions allow learning the observation noise for individual samples as well. This paper is the first to apply these techniques to predictive process monitoring. We found that they contribute towards more accurate predictions and work quickly. However, their main benefit resides with the uncertainty estimates themselves that allow the separation of higher-quality from lower-quality predictions and the building of confidence intervals. This leads to many interesting applications, enables an earlier adoption of prediction systems with smaller datasets and fosters a better cooperation with humans.

\end{abstract}
%

%
\section{Introduction}\label{sec:intro}
Modern information systems and data availability led to the acceleration of process mining research and deployment of its algorithms in industry in recent years. Process mining analyzes event data generated by such information systems with the goal of process discovery, process conformance checking and process enhancement. Predictive process monitoring is an important sub-field of process mining and concerns predicting next events, process outcomes and remaining execution times.
Recent advances in machine learning propelled predictive process monitoring to the next level and many researchers intensified the use of artificial neural networks (NNs) for their predictions.

However, the adoption of these powerful and versatile NNs has not followed suit in practice. Practitioners are reluctant to use NNs that cannot explain their predictions. A related, consequential problem is that NNs are unaware of the uncertainty of their predictions. They will always make a prediction, even when confronted with inputs they were never trained on. This can lead to potentially expensive or even catastrophic mistakes. Uncertainty awareness would therefore be a tremendous asset.

The uncertainty of predictions is the subject of this paper. Our core contribution is the introduction of NN-based uncertainty estimation techniques including heteroscedasticity learning and loss attenuation, concrete dropout and Bayesian neural networks (BNNs) to predictive process monitoring. We test their impact on overall prediction quality, uncertainty estimation quality, and computational time in a carefully designed experimental assessment using three public real-life datasets. Furthermore, we shed light on the practical applications. We consider the problem of remaining execution time prediction of ongoing processes which is highly relevant in practice, as it allows management to stop or alter running processes or initiate other actions. For example, an organization can inform its customers about the expected feedback/fulfillment time for their requests/orders and divert cases with long expected remaining times to a special track to speed them up. 

We define our learning problem and position this paper relative to other work in Section~\ref{sec:def}. Section~\ref{sec:theory} explains two types of uncertainty before introducing techniques adapting plain-vanilla NNs to learn them. We then derive the precise questions we seek to address with our experiments. Section~\ref{sec: eval} describes the setup of these experiments, whose results are presented in Section~\ref{sec:results}. We subsequently present applications enabled by the uncertainty estimates in Section~\ref{sec:working} before summarizing our findings and formulating paths for future research in the final Section~\ref{conclusions}.

\section{Remaining time prediction: definition and related work}\label{sec:def}
In predictive process monitoring, datasets are event logs describing processes, often called \textit{cases}. These cases consist of \textit{events}. A number of attributes, also called \textit{features} or variables, describe these cases and events. In remaining time prediction problems, every event is associated with a \textit{target} feature describing the remaining time until completion of the case.  A \textit{prefix} is an ongoing, incomplete case, with the \textit{prefix length} its number of completed events. Our learning problem is to train a learner using a training dataset containing events, described by their features and organized in prefixes that are labeled with targets, with the goal of predicting the targets of unseen prefixes. \\

In 2008, the first published research on process remaining time prediction \cite{dongen} used non-parametric regressions, followed a few years later by \cite{aalst} proposing to build an annotated transition system. Later, increasingly sophisticated approaches \cite{dataaware} deployed classic machine learning techniques such as support vector regression and naive Bayes and included the events' attributes other than activity name and time into their calculations. Recently, long short-term memory models (LSTMs) entered the scene \cite{awarelstm, tax}. Such deep learning techniques permit the substitution of automatic feature engineering for the error-prone, domain-knowledge-based manual feature engineering of the classic machine learning techniques. The authors of \cite{survey} provide an overview of papers until 2017. Our paper further extends this line of research by complementing the point estimates of these NN with predictions of the respective uncertainty. As such, we realize our goal of not only improving the overall quality of these point estimates, but also of unlocking many applications based on the knowledge of the predictions' uncertainty.

\section{Estimating uncertainty}\label{sec:theory}
\begin{figure}[h]
\vspace*{-.1cm}
\center
\includegraphics[width=1\textwidth]{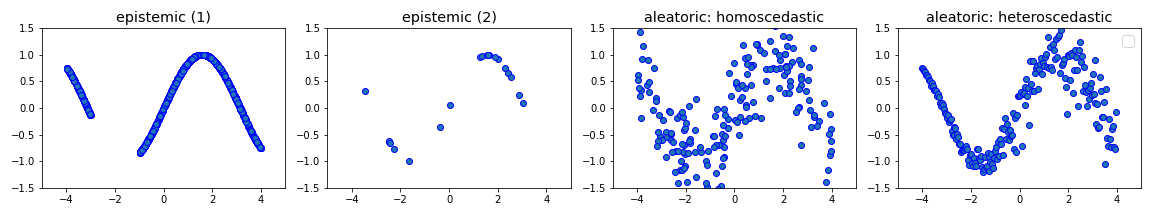}
\vspace*{-.5cm}
\caption{Examples of uncertainty types} \label{fig:uncertainty_types}
\vspace*{-.8cm}
\end{figure}

In the context of predicting with models, we can distinguish two kinds of uncertainty\cite{uncertainty}. The first, the \textit{epistemic} (a.k.a. reducible) uncertainty expresses the model's uncertainty and finds its origin in the paucity of training data. Adding more samples to the training dataset will reduce the epistemic uncertainty. The first two graphs in Fig.~\ref{fig:uncertainty_types} visualize two examples. The second type of uncertainty, the \textit{aleatoric uncertainty} is a measure for the observation noise of the underlying distribution that generated the samples. It is often expressed as $\sigma$ and will not decrease by observing more data. Many models in practice assume the aleatoric noise to be constant  or homoscedastic (as in the third graph in Fig.~\ref{fig:uncertainty_types}). In reality, heteroscedasticity (fourth graph in  Fig.~\ref{fig:uncertainty_types}) is probably much more common: the aleatoric noise varies across the domain.

\subsection{Estimating epistemic uncertainty with Bayesian Neural Networks}
In regular, \textit{deterministic} neural networks, the maximum likelihood estimate (MLE) of a model $\mathcal{H}$'s weights $\bm{\omega}$ maximizes the probability $p(\bm{Y}|\bm{X},\bm{\omega}, \mathcal{H})$ of the observed outcomes $\bm{Y}$ given corresponding inputs $\bm{X}$. Prediction leads to a point estimate $y^*=\mathcal{H}(\bm{x}^*, \bm{\omega})$. Whilst good function approximators, (unregularized) deterministic NNs are prone to overfitting, especially when dealing with small training sets, and therefore struggle dealing with points $\bm{x}^*$ far away from the training data $\bm{X}$. Deterministic models have no knowledge of their point predictions' uncertainty.

The Bayesian approach is \textit{stochastic} by nature: we look for the maximum a posteriori (MAP) distribution of the weights $\bm{\omega}$ given the training set $[\bm{X,Y}]$, that can be expressed using the Bayesian rule:
$$ p(\bm{\omega}|\bm{X},\bm{Y},\mathcal{H}) = \frac{p(\bm{Y}|\bm{X}, \bm{w},\mathcal{H}).p(\bm{\omega}|\mathcal{H})}{p(\bm{Y},\bm{X})}  \text{ or posterior = }\frac{\text{likelihood x prior}}{\text{evidence}}$$
Note that the likelihood equals the MLE problem above. To predict the outcome for a given $\bm{x}^*$, we marginalize the likelihood over $\bm{\omega}$, a process called \textit{inference} ($\mathcal{H}$ dropped to simplify notation):
\begin{equation}\label{eq:inference}
p(y^*|\bm{x^*},\bm{X},\bm{Y}) = \int p(y^*|\bm{x}^*,\bm{\omega}).p(\bm{\omega}|\bm{X},\bm{Y}).d\bm{\omega}
\end{equation}
This is no longer a point estimate, but rather a distribution from which moments (mean, variance, etc.) can be derived. These statistics provide both a point estimate (mean) and a measure of the uncertainty of that estimate (variance), opening a range of possibilities that will be the subject of this paper.
Under certain assumptions, there is an analytical solution to compute the posterior $p(\bm{\omega}|\bm{X},\bm{Y})$ \cite{MacKay} but it is  prohibitively computationally-expensive, as would be Markov Chain Monte Carlo sampling. Consequently, we resort to seeking a closed, approximate function $q_\theta(\bm{\omega})$ over the same domain $\bm{\omega}$ and parameterized by $\theta$. This can be achieved by minimizing the Kullback-Leibler (KL) divergence between the two distributions:
$$\min \, KL\left(q_\theta(\bm{\omega})||p(\bm{\omega}|\bm{X}, \bm{Y})\right)=\int q_\theta(\bm{\omega}). \text{log}\frac{q_\theta(\bm{\omega})}{p(\bm{\omega}|\bm{X}, \bm{Y})}$$
After some mathematical manipulations, the minimization problem above is equivalent to maximizing the \textit{evidence lower bound} (ELBO):
\begin{equation}\label{eq:ELBO}
\text{ELBO}=\mathbf{E}_{q_\theta(\bm{\omega})}\text{log}\, p(\bm{X},\bm{Y}| \bm{\omega})-KL\left(q_\theta(\bm{\omega})||p(\bm{\omega})\right) = \boxed{1}-\boxed{2}
\end{equation}
Maximizing $\boxed{1}$ is the standard MLE approach with $\boxed{2}$ acting as a regularizer keeping the  approximative posterior $q_\theta(\bm{\omega})$ as closely as possible to the prior $p(\bm{\omega})$. Unlike $\boxed{2}$,  the (derivative of) $\boxed{1}$ cannot be computed in closed form. Since the density function $q_\theta(\bm{\omega})$ in $\frac{\partial}{\partial\theta}\int q_\theta(\bm{\omega})\text{log}\, p(\bm{X},\bm{Y}| \bm{\omega}).d\bm{\omega}$ itself depends on $\theta$, regular Monte Carlo (MC) integration is not feasible either. \cite{galthesis} proposes to use the so called \textit{reparameterization trick}\cite{repartrick} to solve $\frac{\partial}{\partial\theta}\int q_\theta(\bm{\omega})\text{log}\, p(\bm{X},\bm{Y}| \bm{\omega}).d\bm{\omega}$. It involves expressing $\bm{\omega}$ as a deterministic function $g(\epsilon,\theta)$ in which $\epsilon$ is a unconditional parameter, allowing to sample $\epsilon$ from $\mathcal{N}(0, I)$ rather than sampling $\bm{\omega}$ from $q_\theta(\bm{\omega})$. The above approach is called \textit{stochastic variational inference}. Often, a Gaussian distribution is placed over every weight $\omega$ in the network with $\omega = g(\epsilon,\theta) = \mu+\sigma.\epsilon$. This method has two serious drawbacks: it doubles the number of parameters to be estimated ($\mu$ and $\sigma$ instead of a single $\omega$ for every node) and requires relatively complex coding.

\textit{Dropout} \cite{hintonDrop} is a popular regularization technique to prevent NNs from overfitting. It resembles training a large number of networks in parallel by dropping out, or randomly ignoring the outputs of nodes (including the network's inputs) during training by multiplying each one by a parameter $\epsilon$ sampled from a Bernoulli distribution with probability $p$. By simply transforming this stochasticity from the feature space in the NNs' dropout scenario to the weight space in BNNs, maximizing ELBO equals minimizing the NNs' dropout loss function with an additional L2 regularizer \cite{galDrop}. We are, thus, able to use standard NNs with easy-to-implement dropout regularization as BNNs, overcoming the drawbacks aforementioned. \textit{Concrete dropout} \cite{concrete} eliminates the need for tuning the dropout parameters $p_i$ (for each layer $i$) by automatically optimizing $p_i$, replacing the discrete Bernoulli distribution with a continuous relaxation (concrete distribution relaxation \cite{concrete2}). In the traditional approach, dropout layers are placed between the convolutional layers in CNNs and only after the inputs and the last LSTM layer in LSTMs. This traditional approach leads to unsatisfying results. In our BNNs (we used both LSTMs and CNNs, see Subsection~\ref{subsec:cnnlstm}), we therefore applied dropout to the inner-product layers (kernels)~\cite{CNN} in CNNs and to all eight weight matrices within the LSTM cells \cite{RNN} which reduces overfitting problems more successfully. 

After training the model as described above, we proceed to inference or prediction by using MC sampling again, performing $T$ stochastic forward passes of our trained model. The predictive mean of Equation~\ref{eq:inference} is estimated by the predictive mean of the MC samples:
\begin{equation}\label{eq:mean}
\mathbf{E}_{p(y^*|\bm{x}^*,\bm{X},\bm{Y})}[y^*]\approx\frac{1}{T}\sum_t\mathcal{H}(\bm{x}^*, \bm{\hat{\omega}})
\end{equation}
with $\bm{\hat{\omega}}$ indirectly sampled from $q_\theta(\bm{\omega})$ by sampling $\epsilon$ from $\mathcal{N}(0, I)$. The variance is given by:
\begin{equation}\label{eq:var}
\text{Var}_{p(y^*|x^*,\bm{X},\bm{Y})}[y^*] \approx\sigma^2+\frac{1}{T}\sum_t\mathcal{H}(\bm{x}^*, \bm{\hat{\omega}})^2-\left( \frac{1}{T}\sum_t\mathcal{H}(\bm{x}^*, \bm{\hat{\omega}}) \right)^2 = \sigma^2 + \boxed{3}
\end{equation}
$\boxed{3}$ is the sample variance of the $T$ stochastic forward passes and can be interpreted as the model's or epistemic uncertainty. Adding more samples to the training dataset will reduce it. Hence, BNNs enable the ability to gauge the model's uncertainty for every prediction made.

\subsection{Estimating heteroscedastic aleatoric uncertainty}
The $\sigma$ in the above Equation~\ref{eq:var} is the aleatoric uncertainty. As most models assume $\sigma$ to be constant, or homoscedastic, over the entire domain, they do not include it in their loss functions (the last term in Equation~\ref{eq:hetero} is simply dropped). However, learning an individual $\sigma_n$ for each sample $n$ would be valuable to better assess the variance of our predictions in Equation~\ref{eq:var}. This is achieved by doubling the last dense layer in the model (unsupervised learning)\cite{uncertainty}. By re-completing the loss function (ignoring the regulation term) to include the learned $\sigma_n$:
\begin{equation}\label{eq:hetero}
L= \min \frac{1}{N} \sum \frac{1}{2\sigma_n^2}(\bm{y}_n-\mathcal{H}(\bm{x}_n))^2 +\frac{1}{2}\text{log}\,\sigma_n^2
\end{equation}
it becomes less sensitive to noisy data, as it will predict high uncertainty for poor predictions and vice versa. This process is called \textit{loss attenuation} and should lead to better overall predictions. The second term in Equation~\ref{eq:hetero} ensures that the model does not simply predict high uncertainty for every sample.

\subsection{LSTM vs. CNN}\label{subsec:cnnlstm}
The techniques described above all depend upon underlying NNs. LSTMs have been the intuitive instrument of choice in predictive process monitoring problems. A growing body of research (e.g. \cite{bai}), however, points at CNNs as a performant and fast alternative for time series and sequence problems, a thesis supported by \cite{cnnlstm} for the related case of process outcome prediction. We, therefore, ran our experiments using both CNNs and LSTMs to gain further insight into the applicability of both models.

\subsection{Objectives}\label{subsec:goals}
Equipped with this understanding, we can now translate our research goal of investigating uncertainty for remaining time prediction into more detailed objectives. First, we assess the effect on the overall quality of point estimates of the following techniques (Subsection~\ref{subsection:overall}):
    \begin{enumerate}
        \item \textbf{Heteroscedasticity}: Estimating the observation loss for individual samples ($\sigma_n$) permits loss attenuation. Can it improve point estimates?
        \item \textbf{Dropout}: BNNs resemble NNs with dropout regularization. What are the merits of isolated dropout in a non-Bayesian context?
        \item \textbf{Concrete dropout}: allows in-model estimating the dropout parameters $p_i$. How does it affect results?
        \item \textbf{BNN}: Using the heteroscedastic NNs with concrete dropout, we apply MC sampling ($T$ stochastic forward passes) and average to calculate point estimates (Equation~\ref{eq:mean}). Do we get better predictions?
        \item \textbf{CNN/LSTM/base case}: We compare CNNs to LSTMs, as well as to a baseline to get an intuition for the absolute performances.
    \end{enumerate}
Second, we look into the uncertainty estimates' ability to separate good from bad predictions and the quality of the confidence intervals based on these uncertainty estimates (Subsection~\ref{subsection:different}). Third (Subsection~\ref{subsec:speed}), we wish to gain insights in the computation time for training and inference respectively. Finally, our fourth objective (Section~\ref{sec:working}) is to explore and assess applications stemming from the knowledge of predictions' uncertainties.

\section{Experimental setup} \label{sec: eval}

Our experimental setup was guided by the goal of answering the research questions in the context of process remaining time prediction, rather than obtaining the highest possible accuracy.

\subsection{Datasets}
 We used three publicly available datasets from the BPI Challenges\footnote{https://data.4tu.nl (4TU Centre for Research Data)}. BPIC\_2017 is a rich and large dataset containing logs of a loan application process at a Dutch bank. BPIC\_2019 is comparable in size but has much shorter cases and concerns a purchase order handling process. BPIC\_2020 is a collection of five smaller datasets related to travel administration at a university. The five subsets are records of processes covering international declaration documents (Intl. Declarations), expense claims (Travel Costs), travel permits (Permits), pre-paid travel costs and requests for payment (Payments) and domestic declaration documents (Domestic Declarations). Our target for all these datasets was defined as the fractional number of days until case completion. 

\subsection{Preprocessing}
To maintain a realistic setting, we refrained from filtering. Other than adding a few synthetic features based on the event time stamps (e.g. event number, elapsed time since previous event, day of the week, ...), we did not apply any domain knowledge whatsoever to our approach. The chronologically 15\% last starting cases (10\% for BPIC\_2020) were withheld as a test dataset . Since the duration of a case is only known at its end (when the process is finished), we deleted all cases from the remaining training set that ended after the start of the first test dataset case\footnote{A theoretical possibility of data leakage remains. In reality, some case variables such as ``Amount'' are possibly unknown at the beginning of the case, even though every event log has a value for them.}. This left us with approximately two thirds of the original cases for BPIC\_2017 and BPIC\_2019. Given the shorter recording time frame for BPIC\_2020, this approach drastically reduced the number of samples for training, especially where cases take longer (Intl. Declarations is only left with 57 events from five cases in the training set). With longer cases (with more deviations) and more levels for the categorical variables, BPIC\_2017 differs significantly from BPIC\_2019. To add further variety, we worked with more features in BPIC\_2017 (10) than in BPIC\_2019 (5). To observe how results depend on the training set size, we performed our experiments on different shares of the available samples for both large datasets, ranging from 0.1\% to 100\%. Table~\ref{tab:data} shows the respective datasets' key statistics and illustrates their differences.\\

\begin{table}
\centering
\vspace*{-1mm}
\caption{Statistics of the used datasets.}\label{tab:data}
\scalebox{0.9}{
\begin{tabular}{|l|c|c|c|c|c|c|c|c|}
\hline
Dataset  & Avg.  & Share of     &Training      &Validation   &Test   &Range &Cate- &Levels\\
                &case       & events               & events           & events             & events & features & gorical &\\
                                & length     & used               &            &              & &  &  features&\\
\hline
BPIC\_2017  & 38.5 &.001 & 629 & 220 & 181,189 & 5 & 5 & 113\\
                            &                                           & .002 & 1,286 & 363& &&&\\
                            &                                             & ...     & ...&... &&&&\\
                            &                                             & .5 & 327,959&79,190 & &&&\\
                            &                                             & 1 & 655,271& 159,306& &&&\\
\hline
BPIC\_2019 &  5.2                                                 & .001 & 625 & 192&162,753&3&2&18\\
                       &                                                 & .002 & 1,263&341 &&&&\\
                       &                                                & ...     &... & ...&&&&\\
                      &                                                  & .5 & 328,994&85,622 &&&&\\
                      &                                                  & 1 & 657,187 & 171,724&&&&\\ 
\hline
\hline
Intl. Declarations&   29.6                                                                    & 1&57 &20 &4,416 & 3&3 & 18\\ 
Travel Costs&  7.7                                                 & 1& 1,706& 412& 1,652& 5 &9 & 74\\
Permits&10.0      & 1& 8,030& 2,132&6,537 & 5&9 &94 \\
Payments& 5.3                                                     & 1& 21,049& 5,743&3,746&4&8&107\\
Domestic Declarations&  8.1                                        & 1&23,434 &6,216 &3,533 & 4& 6& 66\\

\hline
\end{tabular}
}
\vspace*{-1mm}
\end{table}
Range features were standardized. The number of levels of categorical variables was not clipped (non-frequent labels may be a reason for uncertain estimates). The labels were mapped to integers that were then passed to an embedding layer in the neural networks. All possible prefixes were derived from the cases and then standardized to a pre-determined \textit{sequence length} by padding the shorter and truncating the longer ones. All experiments were coded in Python/Pytorch and ran on a desktop with a 3.50 Ghz CPU, 64 Gb of RAM and GeForce 1080 GPU. Our code is published on GitHub\footnote{https://github.com/hansweytjens/uncertainty-remaining\_time} for reproducibility. The metric used was the mean absolute error (MAE). 

\subsection{Estimating the epistemic, aleatoric and total uncertainty}
In the case of BNNs, we performed $T=50$ stochastic forward passes (MC sampling) for every prefix in the test set, each time with a different mask over the weights, by sampling a different $\epsilon$ for every $\omega$ at every run (as per Equation~\ref{eq:mean}). The final predictions are the averages over these 50 samples, discussed in Section~\ref{sec:results}. Using their variance, we calculated the model's uncertainty, i.e. the epistemic uncertainty, for every prediction in the test set using $\boxed{3}$ in Equation~\ref{eq:var}. Moreover, we computed the per-point aleatoric uncertainty in an additional final dense layer in the models and included it in the loss function as in Equation~\ref{eq:hetero}. We added together both types of uncertainty to calculate the total uncertainties used in Subsection~\ref{subsection:different} and Section~\ref{sec:working}. All predictions in the following are averages of 20 runs of the respective models.

\subsection{Base case}
Despite the widespread use of public datasets in predictive process monitoring, assessing the quality of different methods remains hard as the filtering of the datasets, other preprocessing steps, model architectures, etc. are far from uniform across papers. Furthermore, the metrics used allow for comparisons of the methods within a paper but fail to convey an intuition about their absolute merits. To remedy the latter, we included the transition system-based method~\cite{aalst} as a baseline in our experiments.

\section{Results}\label{sec:results}

\subsection{Overall performance}\label{subsection:overall}
We investigated whether the techniques in Subsection~\ref{subsec:goals} contribute to achieving more accurate point estimates. The results are summarized in Fig.~\ref{fig:summary_results} in which every row pertains to a dataset (BPIC\_2017, BPIC\_2019, BPIC\_2020 respectively). Every column compares two or more techniques and will be discussed in the five following subsections. The horizontal axis in the graphs for BPIC\_2017 and BPIC\_2019 represents the share of the available training set that was used for training, ranked from small to large. In the last row, however, it is the five sub-datasets that are ranked from small to large. The vertical axis represents the models' MAE, with the scale being shared throughout the respective rows. Note that we normalized the MAE in the last row, with the respective base cases equal to one. 

\begin{figure}[p]
\center
\includegraphics[width=1\textwidth]{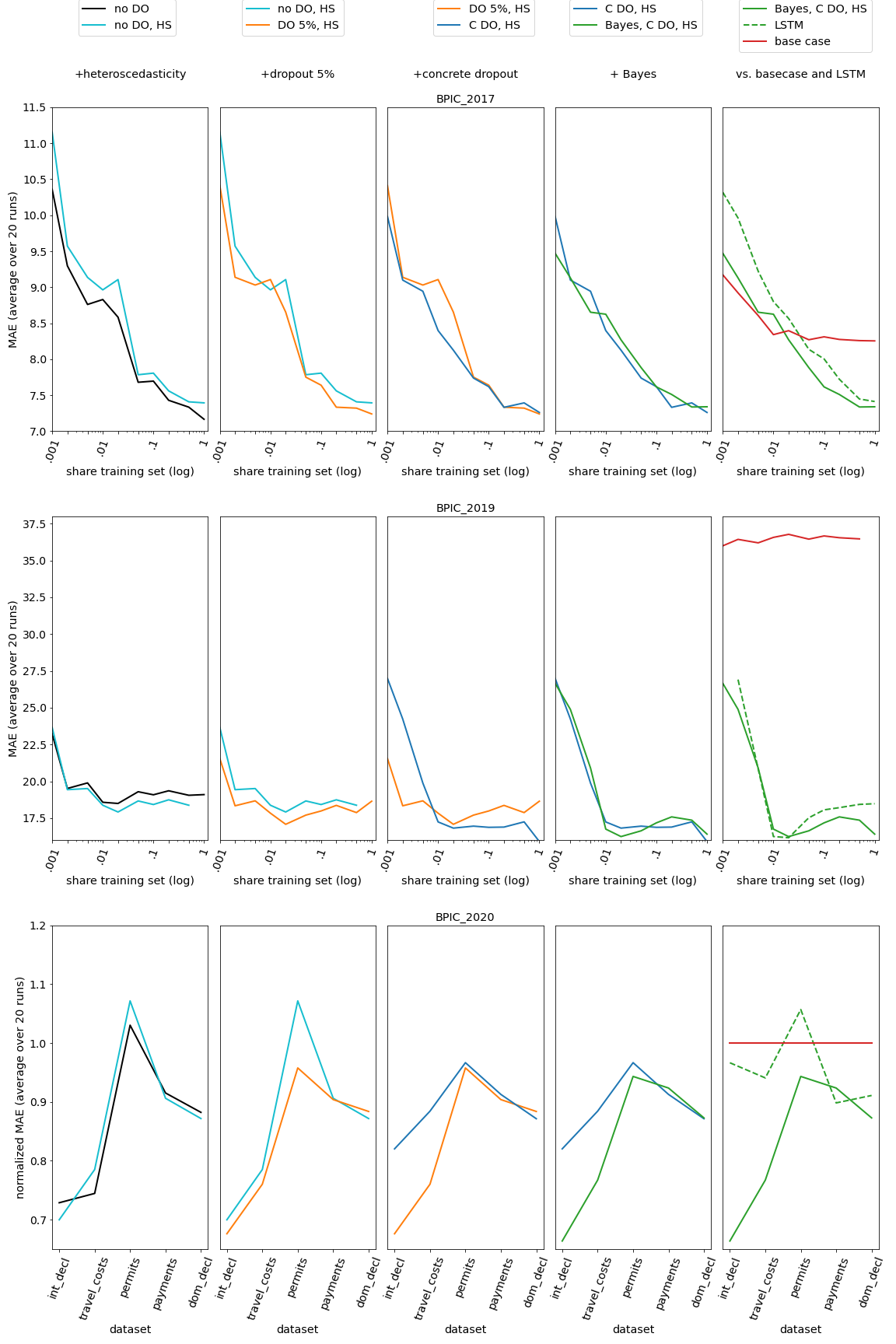}
\caption{Overall results on complete test sets. no DO: no dropout = plain-vanilla NN, HS: heteroscedastic, DO 5\%: 5\% dropout probability, C DO: concrete dropout, Bayes: BNN. Rows show three datasets, stepwise different techniques in columns. BPIC\_2020 results normalized with base case = 1.} \label{fig:summary_results}
\end{figure}

 \vspace*{-4mm}
\subsubsection{Loss attenuation inconclusive (Fig.~\ref{fig:summary_results}: column 1).}
We found no evidence in our experiments for the thesis that learning the heteroscedastic uncertainty and using it by introducing loss attenuation (Equation~\ref{eq:hetero}) in the loss functions leads to more accurate predictions. The black lines in Fig.~\ref{fig:summary_results} represent the plain-vanilla NNs, whereas the cyan lines stand for models including the technique. Results on BPIC\_2017 and BPIC\_2020 significantly worsened. Only in the case of BPIC\_2019 did the technique lower MAE. Two effects could explain that. First, the added complexity may require larger datasets. Second, for datasets with rather homoscedastic aleatoric noise, or datasets with a rather randomly distributed heteroscedastic aleatoric noise, one cannot expect superior results from introducing loss attenuation. We did not further investigate this matter. Nevertheless, learning heteroscedastic uncertainty is indispensable for judging the quality of predictions. We will treat this in Subsection~\ref{subsection:different}.

 \vspace*{-4mm}
\subsubsection{Dropout effectively combats overfitting (Fig.~\ref{fig:summary_results}: column 2).}
The heteroscedastic models are again represented by the cyan lines in column 2. They already included an early-stopping mechanism. But, since the validation sets were in certain cases very small, and some concept drift may exist in the datasets, some overfitting still happened. The dropout mechanism (orange lines) successfully further reduced overfitting on practically all datasets and training set sizes.

 \vspace*{-4mm}
\subsubsection{Concrete dropout works for medium to large datasets (Fig.~\ref{fig:summary_results}: column 3).}
When comparing classic dropout with a fixed dropout parameter (orange lines) to concrete dropout (blue lines), our experiments suggest that, for some very small to small datasets (BPIC\_2019 $<1\%$, BPIC\_2020), concrete dropout negatively affects the overall quality of the predictions. For all other datasets, concrete dropout appeared to work or even improve results. The use of concrete dropout also eliminates the need for the expensive optimization of the dropout parameter(s) $p_{(i)}$ that requires part of the training set to be set aside as a validation set. 

 \vspace*{-4mm}
\subsubsection{Bayesian learning improves results for very small datasets (Fig.~\ref{fig:summary_results}: column 4).}
Until now, we used deterministic NNs to arrive at such models using concrete dropout (blue lines). In column 4, we introduce stochastic NNs in the form of BNNs (green lines), that predict distributions of which the arithmetic averages yield point estimates. BPIC\_2017 and especially BPIC\_2020  support the claim that BNNs produce superior results for smaller datasets. For larger datasets, the effect is negligible, possibly slightly negative. As explained in Section~\ref{sec:theory}, the variance of the produced distributions can be interpreted as a measure for the models' (epistemic) uncertainty, a property we use below. 
As mentioned in Section~\ref{sec:theory}, BNNs by default add L2 regularization to the dropout models. Since the combination of these regularization techniques (in our case even with early-stopping on top) makes these models so robust to overfitting, it is recommended to build models with large capacity to avoid underspecification and train them sufficiently long.

 \vspace*{-4mm}
\subsubsection{CNNs outperform LSTMs, BNNs outperform the base cases (Fig.~\ref{fig:summary_results}: column 5).}
The models in columns  1-4 were all CNNs. When comparing the last one (BNN, full green line) with an otherwise identical LSTM model (dotted green line), it becomes apparent that the CNNs nearly always outperformed the LSTMs. Of course, the chosen architectures (number of layers, nodes, etc.) influenced these outcomes, but the results support similar findings in \cite{bai, cnnlstm}. Unless otherwise mentioned, we will use these heteroscedastic Bayesian CNNs with concrete dropout in the remainder of this paper and simply refer to them as BNN. With the exception of shares of less than 2\% of the BPIC\_2017 dataset and of the BPIC\_2019 Permits dataset, the BNNs outperformed the base cases.

\subsection{Uncertainty estimates}\label{subsection:different}
We analyze the quality of the total uncertainty estimates, focusing on their correlation with the quality of the predictions and on the reliability of confidence intervals based on them.
 \vspace*{-4mm}
\subsubsection{Certainty of predictions correlates strongly with accuracy.}

\begin{figure}[h]
\vspace*{-.1cm}
\center
\includegraphics[width=1\textwidth]{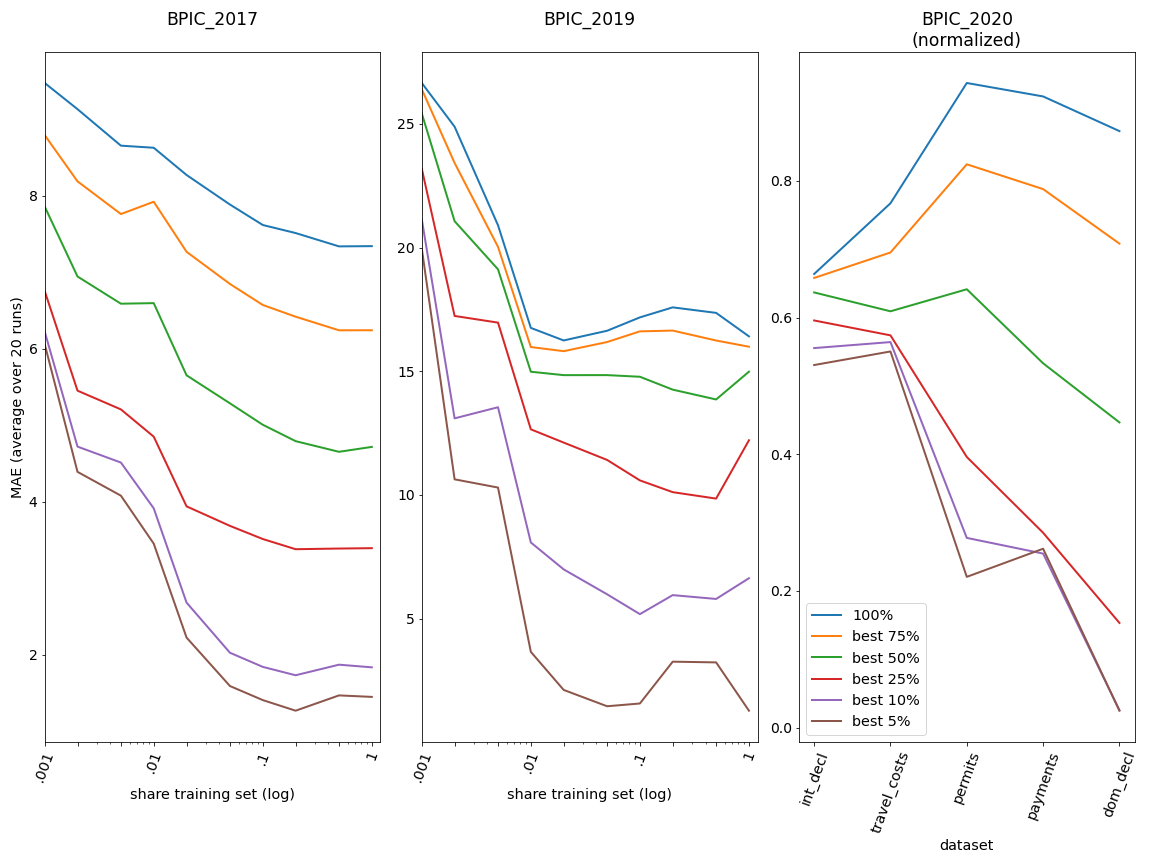}
\vspace*{-.7cm}
\caption{We ranked the samples in the test sets based on the sum of the predicted epistemic and aleatoric uncertainties. In all three datasets, we observe lower MAE (better predictions) for lower levels of uncertainty. We used BNNs with concrete dropout and heteroscedasticity.} \label{fig:uncertainty}
\vspace*{-.3cm}
\end{figure}

We ranked the predictions in the test set and then retained different shares of the predictions while rejecting the others for different uncertainty thresholds (100\%, 75\%, 50\%, 25\%, 10\%, 5\% best). Figure~\ref {fig:uncertainty} shows how well this worked for all datasets and dataset sizes: higher uncertainty led to worse predictions, without fail. Unfortunately, the quality of the uncertainty estimates suffered together with the quality of the predictions when datasets became too small, thus also reducing the possibility to separate good from bad predictions as can be witnessed at the left end of the graphs in Figure~\ref {fig:uncertainty}.

 \vspace*{-4mm}
\subsubsection{Predictions with confidence intervals.}
\begin{figure}[h!]
\hspace*{-1.5cm}   
\includegraphics[width=1.2\textwidth]{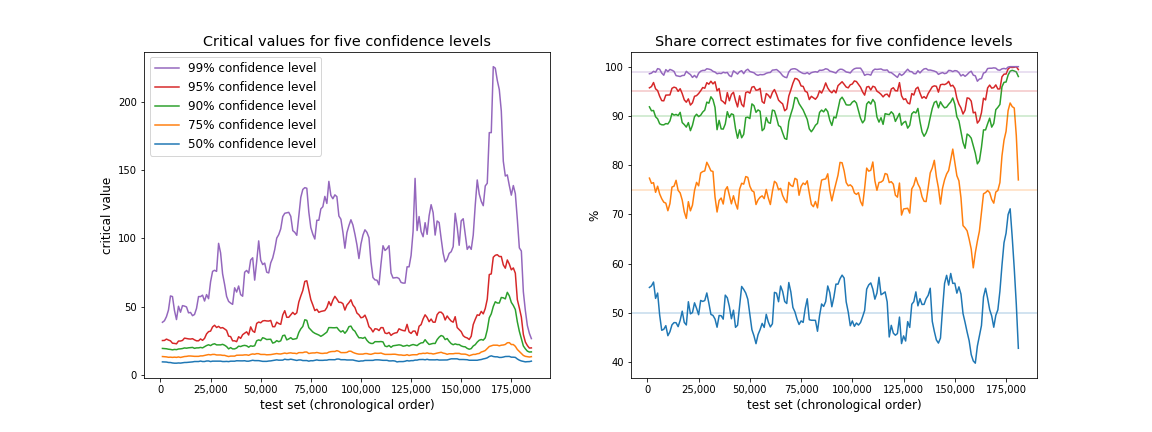}
\caption{Left: critical values for confidence levels of 50\%, 75\%, 90\%, 95\% and 99\% computed on 5,000 preceding samples every 1,000th sample in the test set. Right: Corresponding share of true values in following 5,000 samples within the confidence interval. Dataset is BPIC\_2017 (complete).} \label{fig:calibration}
\end{figure}

To build a confidence interval around a point estimate, the product of a so called \textit{critical value} ($z^*$ in statistics) and the uncertainty is added/subtracted to/from that point estimate to determine the upper/lower bound of the confidence interval. For each desired confidence level (50\%, 75\%, 90\%, 95\%, 99\%) we computed the required critical value based on the last 5,000 samples in the training set. Since the BPIC\_2017 dataset exhibits drift (changes over time), it did not suffice to determine these critical values only once: they had to be calculated online, as can be seen in the left part of Figure~\ref{fig:calibration}. In the right part of Figure~\ref{fig:calibration}, the real shares of true values in the respective confidence intervals are shown. They oscillate around their ideal values (horizontal lines), proving their reliability.

\subsection{Computation time}\label{subsec:speed}

\subsubsection{BNNs train and predict relatively fast.}
To gain an insight in the computation time of BNNs, we disabled the early stopping mechanism and trained the models for 20 epochs on the complete BPIC\_2017 training set. Training the BNNs took around 335 seconds,  approximately 38\% more than the corresponding plain-vanilla deterministic models' 242 seconds. As inference requires MC sampling (we performed 50 MC forward passes), BNN predictions took longer (32 vs 0.65 seconds for all 181,189 test set points). Whilst in most settings the inference time is  low enough to ignore, this may not be the case in certain online environments requiring near-instantaneous decisions. 

Compared to plain-vanilla, deterministic models, the BNNs' hyperparameter space is definitely of a lower dimensionality. There is no need to determine values for the dropout parameter(s) $p_i$ (assuming concrete dropout), model size (we can safely use large-capacity BNNs), number of epochs trained, etc. This may turn their small speed disadvantage into a considerable advantage.

 \vspace*{-4mm}
\subsubsection{CNNs outspeed LSTMs.}
CNNs train nearly an order of magnitude faster than LSTMs that require non-parallelizable sequential calculations. The custom coding to implement dropout within the LSTM cells prevented us from using the very efficient standard PyTorch neural network libraries we used for the CNNs. As a result, our LSTM models slowed down even further and kept us from publishing a fair speed comparison in our specific setting.

\section{Applications of uncertainty}\label{sec:working}
The knowledge of a prediction's quality opens the door to useful practical applications:

 \vspace*{-4mm}
\subsubsection{Higher accuracy and acceptance of prediction systems.}
The previous section demonstrated how the techniques we introduced will generally lead to more accurate overall predictions. However, a yet much higher accuracy can be reached by concentrating on the most certain predictions. An organization requiring a given accuracy threshold can now deploy a prediction system that does not reach that threshold overall but that is aware which of its predictions are expected to surpass it. Predictions that do not reach the (un)certainty threshold can be ignored or passed to humans or another system. In summary, not only can models produce better predictions, but they will also flag potentially incorrect, absurd or even dangerous predictions.

 \vspace*{-4mm}
\subsubsection{Improved human-machine symbiosis.}
The ability to isolate inaccurate predictions permits two-track systems. Cases with good predictions remain on the automated track. Cases with predictions below an uncertainty threshold are passed to the human track. These latter cases will generally be the hardest to solve, more irregular, more interesting ones which could lead to more satisfying work for the involved humans and a better leverage of their cognitive faculties.

 \vspace*{-4mm}
\subsubsection{Working with smaller datasets: earlier adoption of prediction systems.}
As Figures~\ref{fig:summary_results} and~\ref{fig:uncertainty} show, the lack of data often leads to underperforming predictions systems. Organizations will not deploy them or delay their adoption until they feel their dataset is large enough. This may lead to a competitive disadvantage in this digital era requiring rapid innovations, speedy implementation and constant learning where waiting for perfection is no longer an option. The ability to identify predictions that meet a pre-set uncertainty threshold allows for a much faster adoption of prediction systems. Originally, only a relatively small share of the best predictions is actually used. But as the dataset grows, that share continually increases. During this phase-in period, the organization will gain invaluable information to further improve its systems and data collection otherwise lost when remaining on the sidelines.

 \vspace*{-4mm}
\subsubsection{Uncertainty-based analysis.}
\begin{figure}[h]
\vspace*{-1.cm}
\center
\includegraphics[width=.80\textwidth]{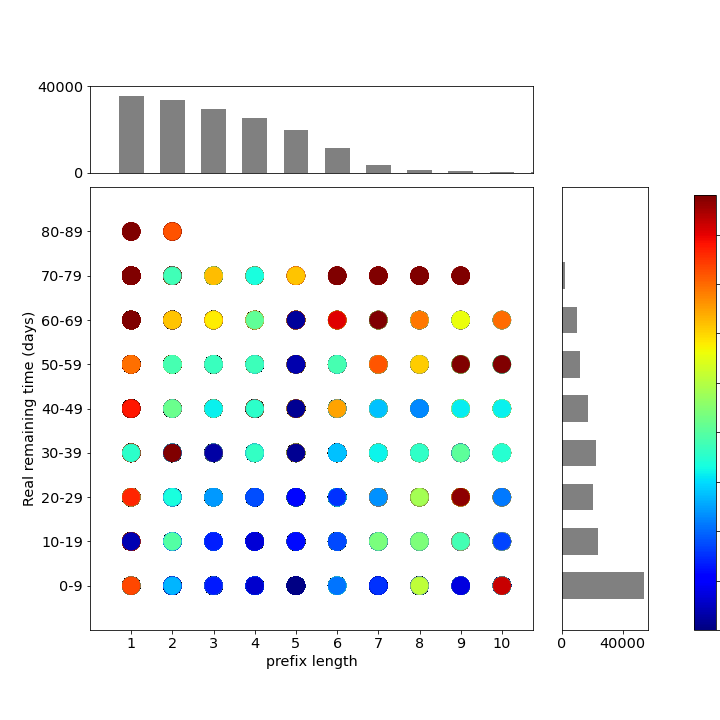}
\vspace*{-.8cm}
\caption{BPIC\_2019, 20\% of training set: Uncertainty (blue=low, red=high) in function of prefix length and real number of remaining days. Grey bars indicate frequency of occurence. Prefix length cut of at 10, corresponding to $>$99\% of samples in test set.} \label{fig:analysis}
\vspace*{-.3cm}
\end{figure}

The estimates of the predictions' uncertainty enables further analysis. For example, as in Fig.~\ref{fig:analysis}, we can plot the test set uncertainty in function of the prefix length and the real number of remaining days (unknown to the model). Given their high aleatoric uncertainty, the model is rightfully very uncertain about the prefixes of length one (first column). The model clearly gains in confidence when prefixes get longer, at least for the most common remaining time lengths (lower than 4 days, lowest four rows). When prefixes start getting longer than six events, the model becomes increasingly wary of its predictions again. Indeed, parts of the domain with fewer samples (e.g. prefix length $>$ six events, real remaining time $>50$ days) should have a higher epistemic, and hence total uncertainty. Outliers, such as the confident predictions of prefixes with length five or those in the second row (10-19 days) of prefix length one, deserve closer attention and may lead to interesting insights. Of course, the uncertainty can be plotted against any other feature as well. A detailed analysis falls outside this paper's scope.

\section{Conclusion and future work}\label{conclusions}
The stochastic Bayesian approach leads neural networks to predict distributions rather than point estimates. These distributions can be used both to derive more precise point estimates (mean) and to estimate the model's epistemic uncertainty (variance). It can be proven that BNNs are nearly identical to deterministic NNs with dropout, which makes them easy to implement. Concrete dropout renders optimizing the dropout parameters $p_i$ obsolete. A dataset's heteroscedastic aleatoric noise can be learned in-model by means of a simple modification to the model and its loss function (loss attenuation).
Whilst inconclusive on the benefits of loss attenuation, this paper shows how dropout, concrete dropout and BNNs generally contribute to more accurate remaining time predictions. CNNs prove to work better and faster than LSTMs. Not all of these techniques work well on all datasets: small datasets pose problems for concrete dropout while they benefit from the Bayesian models that themselves add no value with larger datasets. The presented techniques require little extra coding, learn nearly as fast and are less data-hungry than corresponding regular neural networks. Rather than improving overall accuracy, however, the main benefits of learning uncertainty reside with the new options this knowledge enables. Users can set thresholds to retain those predictions that meet any required accuracy, build confidence intervals around predictions, divide cases between computers and humans in a clever way, adopt prediction models earlier before huge datasets are collected, gain additional insights e.g. in the search for anomalies, etc. We hope that the techniques we proposed help remove some of the barriers that slow down or prevent the adoption of neural networks and could help to extract more value from information systems.

This new field of research can be extended in a variety of ways. First, the validity of our results should be tested on a diverse range of datasets to reach more general conclusions. Also other predictive process monitoring regression and classification problems are logical extensions. Dropout is not the only option to implement variational inference, other methods could be tested as well and may have other characteristics. We also believe that the knowledge of uncertainties can lead to more applications than the ones here presented. As we only concentrated on the total uncertainty, evaluating the respective merits of epistemic and aleatoric uncertainty constitutes another path for future research.

\vspace*{-.1cm}

\end{document}